%% file: main.tex
\documentclass[11pt]{article}
\usepackage[margin=1in]{geometry}
\usepackage{graphicx}
\usepackage{booktabs}
\usepackage{amssymb}
\usepackage{amsmath}
\usepackage{multirow}
\usepackage{multicol}
\usepackage{subcaption}
\usepackage{makecell}
\usepackage{pgffor}
\usepackage[T1]{fontenc}
\usepackage{siunitx}
\usepackage{needspace}
\usepackage{caption}
\usepackage{placeins}
\usepackage{textcomp}
\usepackage{bm}
\usepackage{mdframed}
\usepackage{url}
\usepackage{authblk}
\usepackage{svg}

\title{MARCA: A Checklist-Based Benchmark for Multilingual Web Search.}
\author[1]{Thales Sales Almeida}
\author[1]{Giovana Kerche Bonás}
\author[1]{Ramon Pires}
\author[2]{Celio Larcher}
\author[1]{Hugo Abonizio}
\author[2]{Marcos Piau}
\author[1]{Roseval Malaquias Junior}
\author[1]{Rodrigo Nogueira}
\author[1]{Thiago Laitz}
\affil[1]{Maritaca AI}
\affil[2]{Jusbrasil}

\date{}
\begin{document}

\maketitle
\input{text/0_abstract}

\input{text/1_introduction}
\input{text/2_related_work}

\input{text/3_method}
\input{text/4_results}
\input{text/5_conclusion}

\bibliographystyle{plain} 
\bibliography{main}

\end{document}

%% file: text/0_abstract.tex
\begin{abstract}
Large language models (LLMs) are increasingly used as sources of information, yet their reliability depends on the ability to search the web, select relevant evidence, and synthesize complete answers. While recent benchmarks evaluate web-browsing and agentic tool use, multilingual settings, and Portuguese in particular, remain underexplored.
We present \textsc{MARCA}, a bilingual (English and Portuguese) benchmark for evaluating LLMs on web-based information seeking. \textsc{MARCA} consists of 52 manually authored multi-entity questions, paired with manually validated checklist-style rubrics that explicitly measure answer completeness and correctness. We evaluate 14 models under two interaction settings: a Basic framework with direct web search and scraping, and an Orchestrator framework that enables task decomposition via delegated subagents. To capture stochasticity, each question is executed multiple times and performance is reported with run-level uncertainty.
Across models, we observe large performance differences, find that orchestration often improves coverage, and identify substantial variability in how models transfer from English to Portuguese. The benchmark is available at \url{https://github.com/maritaca-ai/MARCA}
\end{abstract}

%% file: text/1_introduction.tex
\section{Introduction}

Large language models (LLMs) are increasingly used as general-purpose sources of information: users ask for facts, comparisons, and up-to-date summaries, often expecting the system to both retrieve evidence and synthesize a reliable answer. This paradigm depends critically on models' ability to search and read the web effectively and to consolidate information without omissions or hallucinations. 

Recent work has begun to evaluate such capabilities via web-browsing and agentic benchmarks~\cite{nakano2021webgpt,wei2025browsecomp,zhou2023webarena,deng2023mind2web}, but most evaluations remain centered on English. Portuguese, despite being one of the most widely spoken languages globally with over 250 million native speakers, has very limited benchmarks and studies that evaluate web search and evidence-grounded answering capabilities.

We introduce \textsc{MARCA} (\textbf{M}aritaca \textbf{A}I \textbf{R}esearch \textbf{C}hecklist ev\textbf{A}luation), a benchmark for evaluating LLMs' ability to find and verify information on the web in a multilingual setting.
% We introduce \textsc{MARCA} (\textbf{Meaning of acronym redacted for anonymity}), a benchmark for evaluating LLMs' ability to find and verify information on the web in a multilingual setting.
\textsc{MARCA} focuses on multi-entity questions that require systems to gather evidence from multiple webpages and produce structured, complete answers.
Each question is paired with a manually curated checklist rubric, enabling fine-grained evaluation of correctness and coverage.

Beyond language, we study how \emph{interaction design} shapes performance.
We evaluate models in two frameworks: (i) a \emph{Basic} setting where the model directly uses \texttt{web\_search} and \texttt{web\_scrape}, and (ii) an \emph{Orchestrator} setting where the model can delegate sub-questions to subagents that perform web interactions.

Our contributions are:
\begin{itemize}
    \item \textbf{A new bilingual benchmark}: parallel English and Portuguese versions of \textsc{MARCA} with 52 manually authored questions spanning 9 domains for evaluating web-search and evidence-grounded answering.
    \item \textbf{Checklist-based evaluation}: manually authored rubrics that measure both correctness and completeness on questions involving multiple entities.
    \item \textbf{Framework comparison}: an evaluation of Basic tool use versus Orchestrator-style delegation.
\end{itemize}

%% file: text/2_related_work.tex
\section{Related Work}

\subsection{Web Browsing and Search Benchmarks}

Evaluating LLMs as systems that actively search and synthesize information from the web has received growing attention.
WebGPT~\cite{nakano2021webgpt} introduced a browser-assisted QA paradigm where models issue queries, navigate pages, and produce grounded answers with human feedback.
More recently, BrowseComp~\cite{wei2025browsecomp} proposed a benchmark of 1,266 questions designed to require persistent, multi-step browsing to find hard-to-retrieve, entangled information.
Extensions of this line include BrowseComp-ZH~\cite{zhou2025browsecomp_zh}, which targets the Chinese web and reveals that browsing ability can degrade substantially outside English, BrowseComp-Plus~\cite{chen2025browsecomp}, which introduces a fixed corpus for fairer and more reproducible evaluation of deep-research agents, and MM-BrowseComp~\cite{li2025mm}, which extends the paradigm to multimodal queries.

Beyond single-answer retrieval, a broader line of work evaluates LLMs as tool-using agents.
WebArena~\cite{zhou2023webarena} and Mind2Web~\cite{deng2023mind2web} test navigation of real or simulated web interfaces, while WebChoreArena~\cite{miyai2025webchorearena} extends these to tedious, memory- and calculation-intensive tasks.
AgentBench and ToolLLM assess broader tool-calling capabilities across diverse APIs.
These efforts primarily measure task completion rather than the quality and completeness of the information retrieved.

Search Arena~\cite{miroyan2025search} takes a complementary approach, collecting over 24,000 crowdsourced interactions with search-augmented LLMs to analyze human preferences in realistic information-seeking settings.
LiveNewsBench~\cite{zhang2026livenewsbench} introduces a scalable, regularly updated benchmark that automatically generates fresh QA pairs from recent news articles, addressing the risk that static benchmarks conflate memorization with genuine search ability.
On the deep-research front, LiveResearchBench~\cite{wang2025liveresearchbench} evaluates long-form, citation-grounded report generation using rubric-based criteria, while Deep Research Bench~\cite{du2025deepresearch} provides a frozen web snapshot for reproducible evaluation of multi-step research agents.

Despite this rapidly growing landscape, most of these benchmarks are English-centric. Multilingual retrieval resources such as mMARCO~\cite{bonifacio2021mmarco} and MIRACL~\cite{zhang2023miracl} evaluate passage retrieval across languages, but none evaluate end-to-end web-interacting agents.

\subsection{Regional and Linguistic Variation in LLM Performance}

Recent work has shown that LLM capabilities can vary considerably across cultural, regional, and linguistic contexts, motivating language-specific evaluation.
The Blend benchmark~\cite{myung2024blend} measures models' understanding of daily activities across countries, showing that models are better attuned to regions with stronger digital representation.
WorldBench~\cite{moayeri2024worldbench} uses socioeconomic data to evaluate how well models recall information about different countries, finding higher error rates for underrepresented regions such as Africa and the Middle East.
TiEBe~\cite{almeida2025tiebe} evaluates factual recall of notable events by country and observes a strong correlation between a country's GDP and model performance. INCLUDE~\cite{romanou2024include} constructed a large-scale evaluation suite of nearly 200,000 QA pairs from local exam sources across 44 languages, demonstrating significant performance differentials across languages and regional knowledge types, and crucially showing that translating English benchmarks is insufficient because it ignores the cultural and regional knowledge of the environments where multilingual systems would be deployed.
ALM-Bench~\cite{vayani2025all} extends this to the multimodal setting, evaluating language models across 100 languages with culturally grounded visual questions; accuracy drops from 88.4\% (English) to 50.8\% (Amharic) for GPT-4o, revealing a consistent performance gap between high- and low-resource languages.

These disparities also manifest in agentic and web-search settings~\cite{almeida2025ticket,romanou2024include}.
BrowseComp-ZH~\cite{zhou2025browsecomp_zh} found meaningful degradation in web-browsing performance when moving from English to Chinese, demonstrating that search benchmarks designed for English do not transfer straightforwardly to other information ecosystems.

For Portuguese specifically, the evaluation landscape has expanded but remains fragmented.
Structured exams such as the Brazilian National University Entrance Exam (ENEM)~\cite{enem1}, university entrance exams~\cite{almeida2023bluex,santos2025bluexr}, and bar exam evaluations~\cite{oabbench} provide standardized assessments. More recently, PoETa~v2~\cite{almeida2025poeta} introduced a comprehensive benchmark of 44 Portuguese tasks, providing the most extensive evaluation of Portuguese LLMs to date.

Despite this progress, no existing Portuguese benchmark evaluates web-search and evidence-grounded answering capabilities, which is the setting where language-specific retrieval dynamics, content availability, and query formulation strategies can meaningfully affect performance. MARCA addresses this gap by providing a bilingual benchmark that allows performance differences to be directly attributed to language, enabling fine-grained diagnosis of when and why models fail in non-English web interactions.

%% file: text/3_method.tex
\section{Methodology}

This section describes the construction of \textsc{MARCA}, a benchmark designed to evaluate the ability of large language models (LLMs) to \emph{find} information on the web. \textsc{MARCA} focuses on multi-entity information-seeking tasks, in which models must identify, retrieve, and correctly associate information about multiple entities.

\subsection{Dataset construction}

\paragraph{Questions.}
\textsc{MARCA} consists of 52 manually authored questions distributed across various categories. The benchmark focuses on information needs that explicitly require enumerating \emph{multiple entities}, often involving the construction of structured lists where each entity must be paired with the correct attributes. Such questions naturally encourage web-search behavior, including querying multiple sources and integrating evidence across entities.

\paragraph{Rubrics.}
Each question is paired with a manually constructed evaluation checklist. A checklist specifies the set of expected entities and the required fields or attributes for a complete answer. During evaluation, these checklists are used to assess both the correctness and completeness of model outputs. Figure~\ref{fig:example_question} shows an example question with its corresponding checklist.

\begin{figure}[h]
\begin{mdframed}
\small
\textbf{Question (English):} ``Please, make a report about the 82nd Golden Globes, reporting only winners for the Film categories.''

\vspace{0.3em}
\textbf{Checklist (14 items, showing 4):}
\begin{enumerate}
    \item The report must mention the film \emph{The Brutalist} as the winner of the \emph{Best Motion Picture -- Drama} category.
    \item The report must mention the actor \emph{Adrien Brody} as the winner of the \emph{Best Actor in a Motion Picture -- Drama} category.
    \item The report must mention the director \emph{Brady Corbet} as the winner of the \emph{Best Director} category.
    \item The report must mention the film \emph{Emilia P\'{e}rez} as the winner of the \emph{Best Motion Picture -- Musical or Comedy} category.
\end{enumerate}
\vspace{0.3em}
\textbf{Question (Portuguese):} ``Por favor, fa\c{c}a um relat\'{o}rio sobre o 82\textordmasculine\ Golden Globes, mencionando apenas os vencedores das categorias de Cinema.''
\end{mdframed}
\caption{Example \textsc{MARCA} question with a subset of its checklist. The Portuguese version is a parallel translation; its checklist mirrors the English one.}
\label{fig:example_question}
\end{figure}

A subset of questions also uses dynamic checklists, where expected values are retrieved at evaluation time. For example, given the question "How many citations does the second author of the Sabia-2 paper have on Google Scholar?", we programmatically scrape the relevant author's page and incorporate the retrieved information into the rubric. This ensures the evaluation reflects the current state of the web rather than a static snapshot.

\textsc{MARCA} is released in parallel English (EN) and Portuguese (PT) versions. This setup allows us to evaluate whether systems exhibit performance degradation solely due to a change in the input query language.

For each language, we also provide language-specific versions of the evaluation checklists. Queries in a given language are always evaluated using the corresponding checklist.

\subsection{Evaluation frameworks}

We evaluate models under two web-search interaction frameworks that reveal different patterns of tool use and information aggregation.

\subsubsection{Basic framework}

In the \emph{basic} framework, the model has direct access to two tools:
\begin{itemize}
    \item \texttt{web\_search}: submits a search query to the Serper API and returns a ranked list of results, including links, titles, and snippets;
    \item \texttt{web\_scrape}: retrieves and returns the cleaned markdown content of a specified URL via the Serper scrape endpoint.
\end{itemize}

In this framework, a single model instance is responsible for the entire information-seeking process. The model issues all queries, scrapes relevant pages, and retains all retrieved content in its context in order to produce a final answer. The model is allowed up to 40 tool-call iterations per question.

\subsubsection{Orchestrator framework}

In the \emph{orchestrator} framework, the system consists of two components: a top-level model (the \emph{orchestrator}) and one or more \emph{subagents}. The orchestrator is provided with a delegation tool, \texttt{delegate\_to\_subagent}, while each subagent has access to the same \texttt{web\_search} and \texttt{web\_scrape} tools available in the basic framework.

This design encourages task decomposition. The orchestrator can split a complex, multi-entity question into smaller sub-questions and delegate targeted evidence-gathering to subagents. Importantly, the orchestrator does \emph{not} interact with the web directly. Instead, it only receives the subagents' outputs, and it only observes page content when that content is explicitly surfaced through excerpts or citations returned by a subagent.

\begin{figure}[h]
    \centering
    \includegraphics[width=0.8\textwidth]{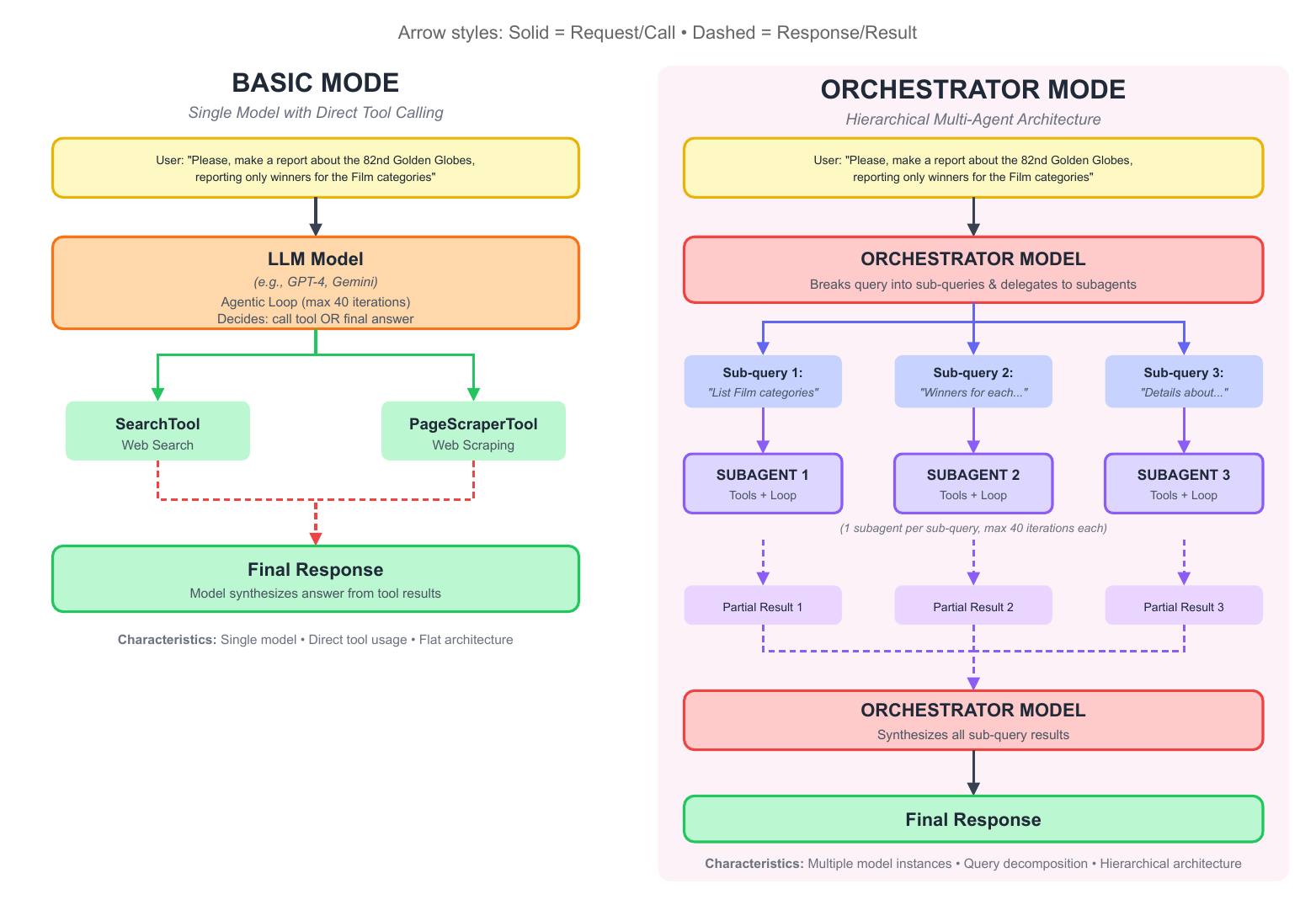}
    \caption{Overview of the two evaluation frameworks. In the \emph{basic} framework, the model directly issues web searches and scrapes pages. In the \emph{orchestrator} framework, a top-level model delegates sub-questions to subagents that perform the web interactions.}
    \label{fig:orchestration_schema}
\end{figure}

\subsection{Evaluation protocol}

For each question, we evaluate the model under a fixed framework (basic or orchestrator) and language (EN or PT). Each question is answered three times ($k=3$) to account for model stochasticity. Every generated answer is then graded against the corresponding checklist by an LLM-based judge.

We use GPT-4.1\footnote{Model version: \texttt{gpt-4.1-2025-04-14}.} as the evaluation judge. The judge receives the original question, the model's answer, and the checklist, and is instructed to apply strict evaluation criteria. For each checklist item, the judge produces a binary decision (satisfied or not satisfied) along with a brief explanation. We use structured output parsing to ensure consistent formatting of judge responses.

\subsubsection{Checklist-based scoring}

Given a checklist with $T$ requirements, the judge produces a binary decision for each requirement (satisfied or not satisfied). Let $C$ denote the number of satisfied requirements.

\paragraph{Checklist Accuracy.}
We define \texttt{checklist\_accuracy} as the fraction of checklist items that are satisfied:
\[
\texttt{Checklist Accuracy} = \frac{C}{T}.
\]

% \paragraph{Threshold metrics.}
% In addition, we report two threshold-based success metrics derived from \texttt{checklist\_accuracy}:
% \begin{itemize}
%     \item \texttt{is\_75\_percent\_correct}: equals 1 if \texttt{checklist\_accuracy} $\ge 0.75$, and 0 otherwise;
%     \item \texttt{is\_100\_percent\_correct}: equals 1 if \texttt{checklist\_accuracy} $= 1.0$, and 0 otherwise.
% \end{itemize}

% These boolean metrics take values in $\{0,1\}$ and are averaged across questions and runs. The resulting mean corresponds to the proportion of cases that meet each threshold.

\subsubsection{Aggregation and uncertainty quantification}

To report aggregate performance together with uncertainty estimates, we adopt a run-level aggregation procedure. For a metric $M$ and $N$ questions, let $M[q,r]$ denote the score for question $q$ in run $r$, where $r \in \{1,\dots,k\}$.

\paragraph{Run-level averages.}
For each run $r$, we compute the average score across all questions:
\[
\texttt{run\_avg}[r] = \frac{1}{N}\sum_{q=1}^{N} M[q,r].
\]

\paragraph{Mean and error bars.}
We report the overall mean as the average of the $k$ run-level averages,
\[
\mu = \frac{1}{k}\sum_{r=1}^{k} \texttt{run\_avg}[r],
\]
and quantify uncertainty using the sample standard deviation across runs, with Bessel's correction ($\mathrm{ddof}=1$):
\[
\sigma = \sqrt{\frac{1}{k-1}\sum_{r=1}^{k}\bigl(\texttt{run\_avg}[r]-\mu\bigr)^2}.
\]
Results are reported as $\mu \pm \sigma$, and $\pm\sigma$ is used as error bars in plots.
    

%% file: text/4_results.tex
\section{Results}

\begin{table}[t]
\centering
\small
\setlength{\tabcolsep}{6pt}
\begin{tabular}{lcccc}
\toprule
\textbf{Model}
& \multicolumn{2}{c}{\textbf{English}}
& \multicolumn{2}{c}{\textbf{Portuguese}} \\
\cmidrule(lr){2-3} \cmidrule(lr){4-5}
& \textbf{Basic} & \textbf{Orchestrator}
& \textbf{Basic} & \textbf{Orchestrator} \\
\midrule
GPT-5.2~\cite{openai2025gpt52}              & $0.837 \pm 0.062$ & $0.881 \pm 0.039$ & $0.855 \pm 0.036$ & $0.872 \pm 0.022$ \\
GPT-5-mini~\cite{openai2025gpt5blog}           & $0.837 \pm 0.015$ & $0.827 \pm 0.027$ & $0.852 \pm 0.022$ & $0.819 \pm 0.083$ \\
Gemini-3-pro~\cite{google2025gemini}         & $0.885 \pm 0.020$ & $0.897 \pm 0.042$ & $0.896 \pm 0.017$ & $0.889 \pm 0.016$ \\
Gemini-3-flash~\cite{google2025gemini}       & $0.828 \pm 0.008$ & $0.831 \pm 0.013$ & $0.733 \pm 0.016$ & $0.820 \pm 0.005$ \\
GPT-4.1~\cite{openai2025gpt41}              & $0.815 \pm 0.022$ & $0.779 \pm 0.026$ & $0.826 \pm 0.023$ & $0.825 \pm 0.029$ \\
GPT-4.1-mini~\cite{openai2025gpt41}         & $0.640 \pm 0.023$ & $0.719 \pm 0.027$ & $0.631 \pm 0.014$ & $0.735 \pm 0.045$ \\
Sabia-4~\cite{laitz2026sabi}              & $0.743 \pm 0.018$ & $0.832 \pm 0.050$ & $0.695 \pm 0.024$ & $0.800 \pm 0.030$ \\
Sabiazinho-4\cite{laitz2026sabi}         & $0.709 \pm 0.010$ & $0.721 \pm 0.024$ & $0.705 \pm 0.058$ & $0.753 \pm 0.032$ \\
Qwen-3-30B~\cite{yang2025qwen3}           & $0.469 \pm 0.054$ & $0.520 \pm 0.040$ & $0.484 \pm 0.029$ & $0.493 \pm 0.015$ \\
Qwen-3-235B~\cite{yang2025qwen3}          & $0.668 \pm 0.020$ & $0.753 \pm 0.013$ & $0.677 \pm 0.032$ & $0.678 \pm 0.027$ \\
Kimi-K2~\cite{team2025kimi}              & $0.769 \pm 0.035$ & $0.830 \pm 0.026$ & $0.797 \pm 0.039$ & $0.773 \pm 0.049$ \\
Gemini-2.5-Pro~\cite{comanici2025gemini}       & $0.658 \pm 0.066$ & $0.701 \pm 0.025$ & $0.585 \pm 0.076$ & $0.598 \pm 0.037$ \\
Gemini-2.5-Flash~\cite{comanici2025gemini}     & $0.528 \pm 0.034$ & $0.540 \pm 0.019$ & $0.475 \pm 0.072$ & $0.517 \pm 0.018$ \\
Gemini-2.5-Flash-Lite~\cite{comanici2025gemini} & $0.260 \pm 0.015$ & $0.317 \pm 0.036$ & $0.299 \pm 0.068$ & $0.354 \pm 0.020$ \\
\bottomrule
\end{tabular}
\caption{\texttt{Checklist Accuracy} (mean $\pm$ run-level standard deviation) across languages and inference frameworks.}
\label{tab:checklist_accuracy_lang_framework}
\end{table}

Table~\ref{tab:checklist_accuracy_lang_framework} reports \texttt{Checklist Accuracy} (mean $\pm$ run-level standard deviation) across languages and inference frameworks for 14 models. Overall performance spans a wide range. The strongest systems approach $\approx 0.90$ checklist coverage in their best configurations (e.g., Gemini-3-pro and GPT-5.2), while the weakest setting in the table (Gemini-2.5-Flash-Lite, English Basic) achieves only $0.260$. This spread highlights substantial variation in models' ability to reliably enumerate and attribute multiple entities in web-based information-seeking tasks.

Across models, run-level uncertainty is typically small relative to mean performance for many systems (e.g., GPT-4.1 and Qwen-3-235B), indicating stable behavior across repeated runs. In contrast, some models exhibit noticeably larger variability (e.g., Gemini-2.5-Pro with $\sigma$ up to $0.076$), suggesting greater sensitivity to stochasticity during web interaction and evidence aggregation.

\subsection{Impact of the inference framework}

\begin{figure}[h]
    \centering
    \includegraphics[width=0.8\textwidth]{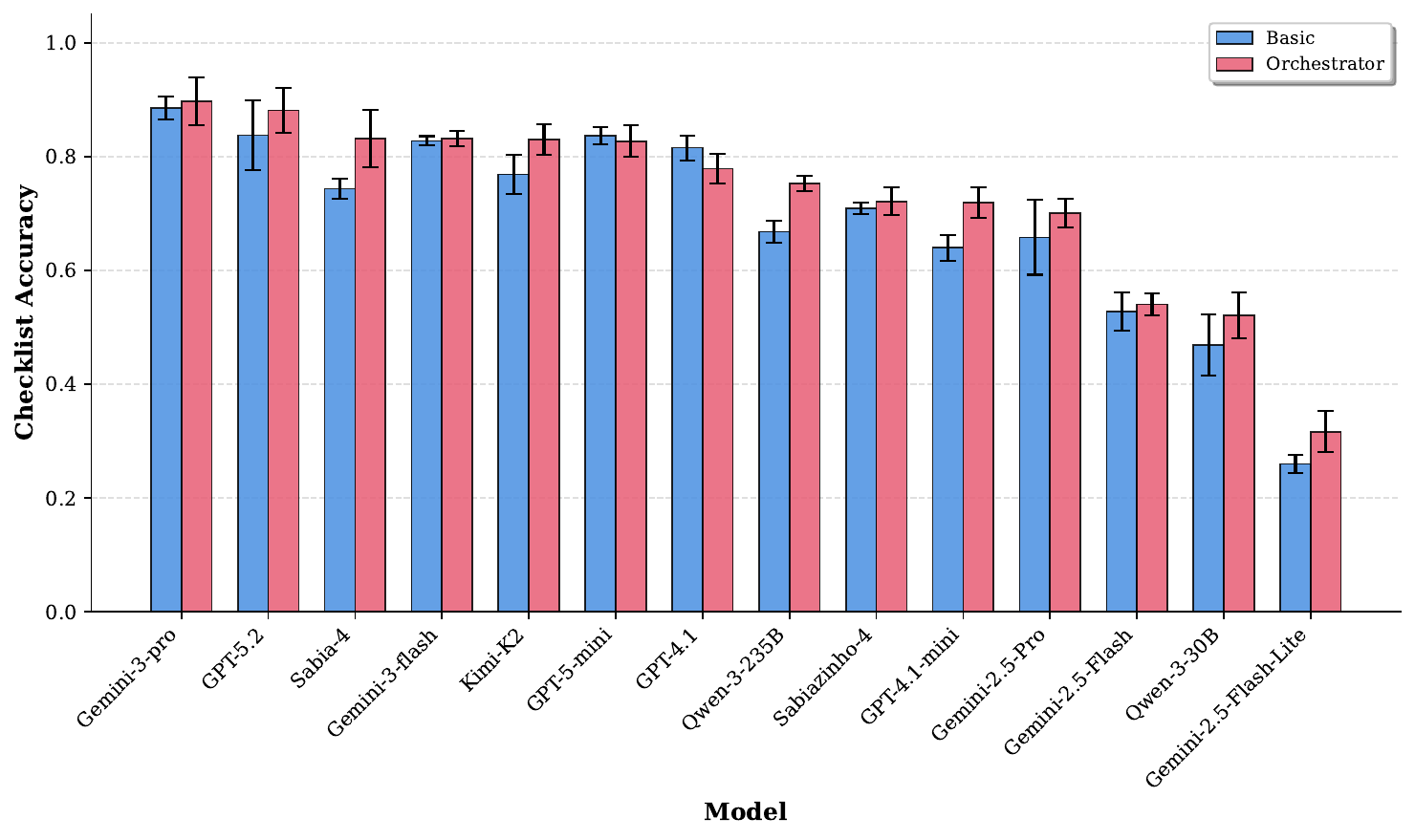}
    \caption{English \texttt{Checklist Accuracy} under Basic vs.
    Orchestrator. Error bars correspond to run-level standard deviation.}
    \label{fig:basic_vs_orchestrator_en}
\end{figure}

Figure~\ref{fig:basic_vs_orchestrator_en} compares English performance under the Basic and Orchestrator frameworks. For most models, the Orchestrator setting yields higher \texttt{Checklist Accuracy} than the Basic setting, indicating that explicit task decomposition and delegation generally improve performance on multi-entity information-seeking queries.

The magnitude of these gains varies substantially across models. Several systems exhibit only modest improvements (e.g., Qwen-3-30B: $0.469 \rightarrow 0.520$; Gemini-2.5-Flash: $0.528 \rightarrow 0.540$), whereas others benefit much more strongly from orchestration (e.g., GPT-4.1-mini: $0.640 \rightarrow 0.719$; Sabia-4: $0.743 \rightarrow 0.832$). This pattern suggests that the Orchestrator framework is particularly effective for models that struggle to manage multi-entity retrieval and consolidation within a single reasoning context. For these models, decomposing the question into focused sub-queries reduces the burden of tracking multiple entities simultaneously, allowing each subagent to concentrate on retrieving and verifying a smaller set of facts.

The effect, however, is not uniform. A small number of strong models show comparable or slightly better performance in the Basic setting, most notably GPT-4.1 ($0.815$ vs.\ $0.779$). One explanation is that the overhead of decomposition---potential miscommunication between orchestrator and subagents, or loss of global context---can outweigh the benefits for models that already handle complex queries effectively.

\subsection{Impact of language}

\begin{figure}[h]
    \centering
    \includegraphics[width=0.8\textwidth]{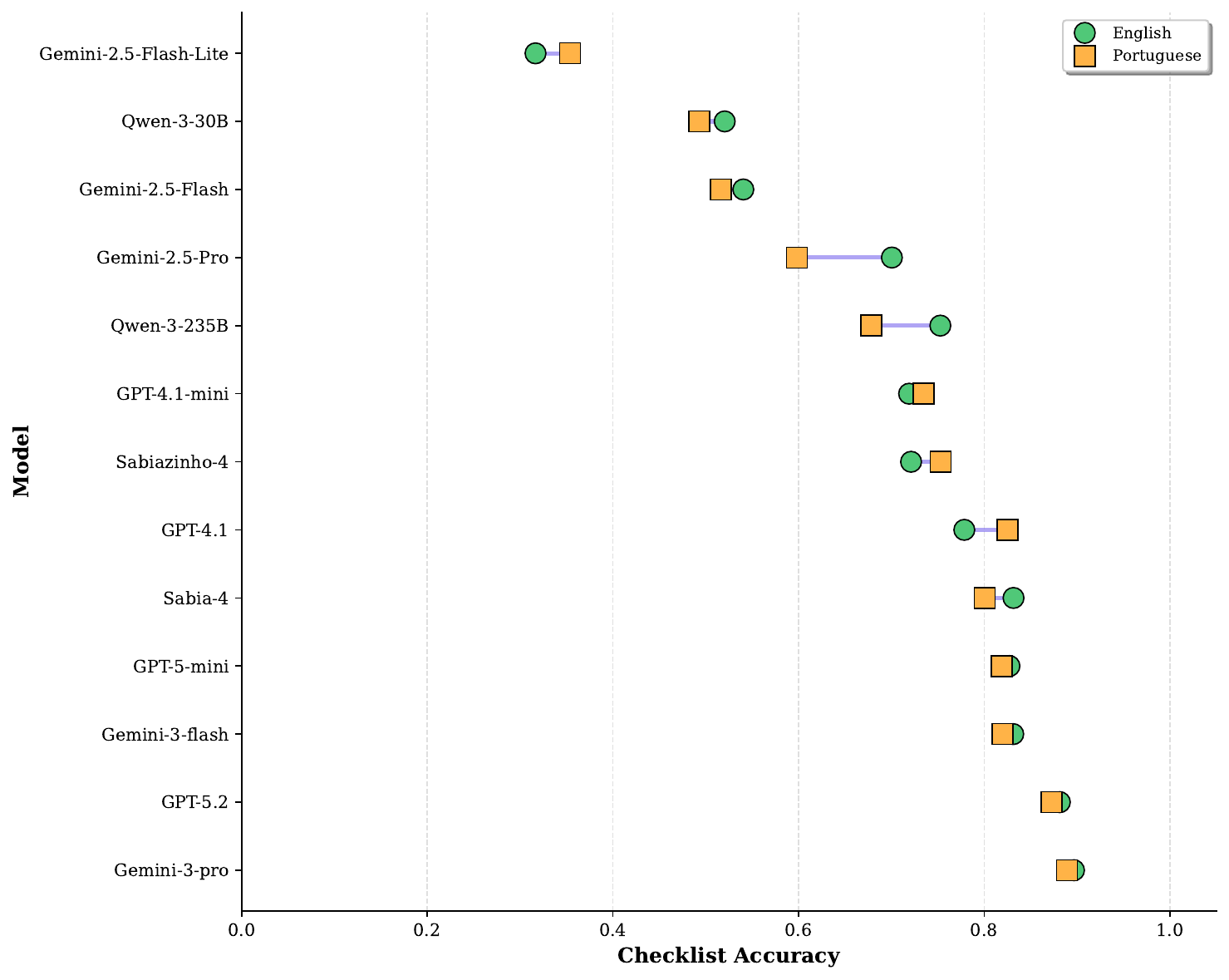}
    \caption{Dumbbell plot comparing \texttt{Checklist Accuracy} between English and Portuguese in the Orchestrator setting (mean with run-level uncertainty).}
    \label{fig:en_vs_pt_orchestrator}
\end{figure}

Figure~\ref{fig:en_vs_pt_orchestrator} compares English and Portuguese performance on \textsc{MARCA}. Overall, results indicate that models are generally capable of operating across languages, but the impact of switching from English to Portuguese varies substantially by model.

Several systems exhibit closely matched performance between the two languages, and in some cases achieve slightly higher checklist accuracy in Portuguese (e.g., GPT-4.1, GPT-4.1-mini, and Sabiazinho-4). In contrast, other models show noticeable degradations when evaluated in Portuguese, including Qwen-3-235B and Gemini-2.5-Pro, with gaps that exceed 5\% in checklist accuracy. The direction and magnitude of these differences are not consistent across model families, indicating heterogeneous multilingual behavior.

This variability suggests that multilingual web information seeking depends on more than surface-level language understanding. Models differ in their ability to formulate effective non-English queries, interpret foreign-language webpages, and consolidate evidence across sources without omitting entities. Notably, some models may implicitly translate Portuguese queries into English for search, which can be effective when English sources are sufficient but fails when the answer requires Portuguese-specific content (e.g., Brazilian government positions or Copa Libertadores details). These results highlight the importance of evaluating multilingual capabilities in realistic web-based settings, where language-specific retrieval dynamics and evidence availability can meaningfully affect performance.

\subsection{Performance and cost trade-offs}

\begin{figure}[h]
    \centering
    \includegraphics[width=0.8\textwidth]{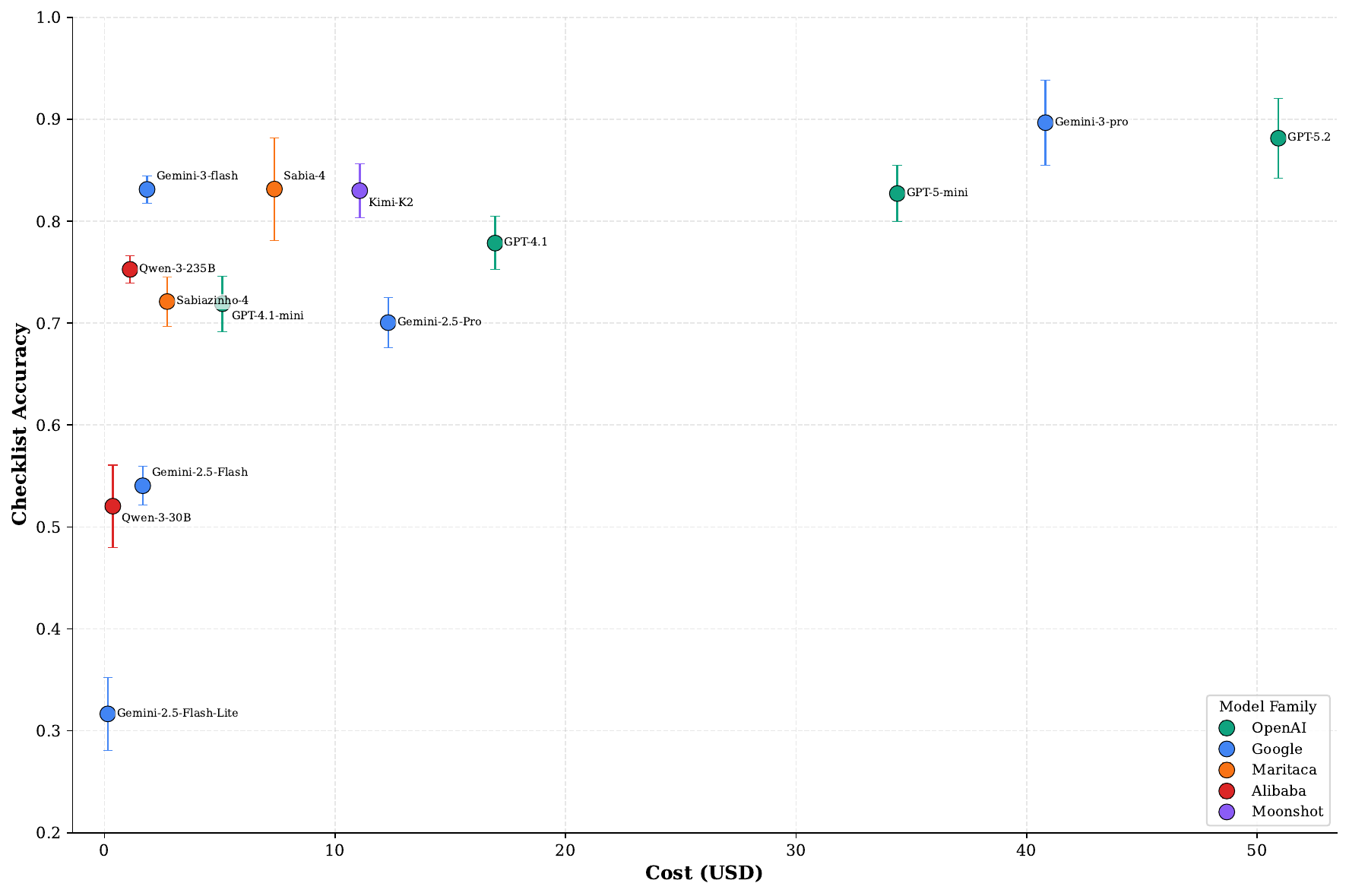}
    \caption{Performance vs.\ cost in English Orchestrator using \texttt{Checklist Accuracy} (mean with run-level uncertainty). Cost is measured as total token usage per question.}
    \label{fig:perf_cost_en_orchestrator}
\end{figure}

Figure~\ref{fig:perf_cost_en_orchestrator} shows the relationship between \texttt{Checklist Accuracy} and inference cost for English queries. Overall, higher checklist coverage tends to come at higher cost, but the relationship is uneven: models with similar accuracy can differ substantially in inference cost.

Lower-cost models generally achieve limited checklist coverage, while higher-performing models reach accuracies close to $0.90$ at significantly increased cost. Between these extremes, several models occupy an intermediate region, offering moderate performance improvements at relatively modest additional cost. This suggests that practitioners can make informed choices based on their accuracy requirements and budget constraints, as the Pareto frontier is not densely populated---there are clear opportunities to select models that offer favorable accuracy-per-cost ratios.

%% file: text/5_conclusion.tex
\section{Conclusion}

We introduced \textsc{MARCA}, a bilingual benchmark for evaluating large language models on web-based information seeking with an explicit focus on multi-entity queries. By pairing manually authored questions with checklist-style rubrics, \textsc{MARCA} enables fine-grained measurement of both correctness and completeness across 9 thematic domains.

Our experiments reveal substantial variation across models, languages, and interaction designs. While orchestration and task decomposition often improve checklist coverage---particularly for models that struggle with multi-entity consolidation in a single context---the gains are not uniform, and some strong models perform competitively without delegation. Similarly, performance transfer from English to Portuguese varies across models, highlighting that multilingual web interaction depends not only on language understanding but also on retrieval behavior and evidence consolidation.

Finally, our analysis of performance--cost trade-offs shows that higher checklist coverage generally comes at increased inference cost, but with significant variation across models in their accuracy-per-cost ratios, providing actionable guidance for practitioners.